\documentclass[conference]{IEEEtran}
\IEEEoverridecommandlockouts
\usepackage{cite}
\usepackage{amsmath,amssymb,amsfonts}
\usepackage{algorithmic}
\usepackage{graphicx}
\usepackage{textcomp}
\usepackage{xcolor}
\usepackage[font=small]{caption}
\def\BibTeX{{\rm B\kern-.05em{\sc i\kern-.025em b}\kern-.08em
    T\kern-.1667em\lower.7ex\hbox{E}\kern-.125emX}}
\usepackage{epsfig,endnotes, graphicx, tikz, pgfplots, csvsimple, array, multirow, subfig, caption}

\begin{document}
%\author{Rashik Shadman, Alvaro Velasquez, Sarwar Murshed, Edward Verenich, Faraz Hussain}
\title{The Utility of Feature Reuse: Transfer Learning in
Data-Starved Regimes
}

\author{\IEEEauthorblockN{Rashik Shadman}
\IEEEauthorblockA{\textit{Electrical and Computer Engineering Department} \\
\textit{Clarkson University}\\
Potsdam, NY, USA \\
\texttt{shadmar@clarkson.edu}}
\and
\IEEEauthorblockN{M. G. Sarwar Murshed}
\IEEEauthorblockA{\textit{Computer Science Department} \\
\textit{University of Wisconsin-Green Bay}\\
Green Bay, WI, USA \\
\texttt{murshedm@uwgb.edu}}
\and
\IEEEauthorblockN{Edward Verenich}
\IEEEauthorblockA{%\textit{dept. name of organization (of Aff.)} \\
\textit{Air Force Research Lab}\\
Rome, NY, USA \\
\texttt{edward.verenich.2@us.af.mil}}

\and
\IEEEauthorblockN{Alvaro Velasquez}
\IEEEauthorblockA{\textit{Computer Science Department} \\
\textit{University of Colorado Boulder}\\
Boulder, CO, USA \\
\texttt{alvaro.velasquez@colorado.edu}}
\and
\IEEEauthorblockN{Faraz Hussain}
\IEEEauthorblockA{\textit{Electrical and Computer Engineering Department} \\
\textit{Clarkson University}\\
Potsdam, NY, USA \\
\texttt{fhussain@clarkson.edu}}
}

\maketitle

\begin{abstract}
The use of transfer learning with deep neural networks has increasingly become widespread for
deploying well-tested computer vision systems to newer domains, especially those with limited datasets. %improve performance of computer vision systems.
%% Its effectiveness largely depends on the nature of the source and target domains
%% as well as the data regime that the target application resides in.
We describe a transfer learning use case for a domain with a data-starved regime, having fewer
than 100 labeled target samples.
We evaluate the effectiveness of convolutional feature extraction and fine-tuning of overparameterized models
with respect to the size of target training data, as well as their generalization performance on data with covariate shift,
or out-of-distribution (OOD) data.
Our experiments demonstrate that both overparameterization and feature reuse contribute to the successful application of transfer learning in training image classifiers in data-starved regimes.
%\todo{Remove this line} The results show that transfer learning technique improves the performance of CNN architectures in data-starved regimes which is confirmed by visual explanation.
We provide visual explanations to support our findings and conclude that transfer learning enhances the performance of CNN architectures in data-starved regimes.
\end{abstract}


\section{Introduction}\label{sec:intro}
Transfer learning (TL) has become an indispensable technique for deploying deep learning assisted computer vision
systems to new domains.
The basic approach is to use existing neural network architectures that were trained on
large natural image datasets such as \textit{ImageNet} \cite{IMAGENet} 
%or \textit{CIFAR} \cite{CIFAR} 
and fine-tune all or some of their weights towards some new applications~\cite{DBLP:journals/corr/SimonyanZ14a, DenseNet:8099726, RESNET:7780459}.
This approach has been successfully applied in medical applications such as
radiology~\cite{ChexNet:DBLP:journals/corr/abs-1711-05225} and ophthalmology~\cite{OPH:10.1167/iovs.16-19964}, image classification\cite{duan2012learning}, software defect classification\cite{nam2015heterogeneous} etc.

A recent study by Raghu et al investigated the effects of TL on medical imaging and concluded that TL from natural image datasets to the medical domain offers limited performance gains
with meaningful feature reuse concentrated at the lowest layers of the networks~\cite{raghu2019transfusion}.
Related to our operational setting was the observation that the benefit of TL from ImageNet-based models to medical models in very small data regimes (which the authors consider those with datasets of 5000 data points or less)
was largely due to architecture size.
%size \todo{how is benefit related to arch size?}
It follows that
overparameterization\footnote{An overparameterized regime refers to a setting where the number of model parameters
exceeds the number of training examples~\cite{oymak19a, NIPS2019_8479}.}
was likely the source of performance gain.
%and the regime's benefit for model convergence is reported in ~\cite{oymak19a, NIPS2019_8479}.

In operational settings, we are often required to create image classifiers for novel classes of objects that are not represented in natural image datasets, and because of the dynamic nature of our applications, our \emph{data-starved regime} only allows for roughly 100 data points per class (i.e., an order of magnitude less than what Raghu et al consider very small data regimes~\cite{raghu2019transfusion}). 

 In this paper, we evaluate the effectiveness of transfer learning and overparameterization of ImageNet-based models towards extremely small operational data. Our hypothesis is that \emph{transfer learning with overparameterized models enables building
   useful image classifiers in operational data-starved environments}, and that
 \emph{feature reuse is a significant enabling mechanism of this transfer}.

 To examine the relationship of feature reuse and overparameterization,
 we trained several image classifiers by fine-tuning a number of pre-trained architectures to recognize a novel class of images.
 Our experiments show that overparameterization aids in useful transfer,
 but the benefit of model size in terms of the number of trainable parameters to model performance levels off as the size of the model grows.
 
 We also show that feature reuse provides significant benefits to learner performance by demonstrating that random initialization of the same architectures in data-starved regimes results in poor generalization compared to models that leveraged the feature reuse of transfer learning for model fine-tuning.
 
Explainability provides the reason behind the prediction/classification/decision of AI systems. Due to the large size and complexity of deep learning models, interpreting their results remains a significant challenge. Explainability assures much needed transparency for deep learning models. In order to justify the effectiveness of transfer learning implemented on ImageNet based CNN models in small data regimes, we provide visual explanations of the decisions of the model. Our visual explanations strengthen
the predictions by the CNN model.
 
\begin{figure} [ht]
\center
\fbox{\includegraphics[width=0.9\linewidth, height=14cm]{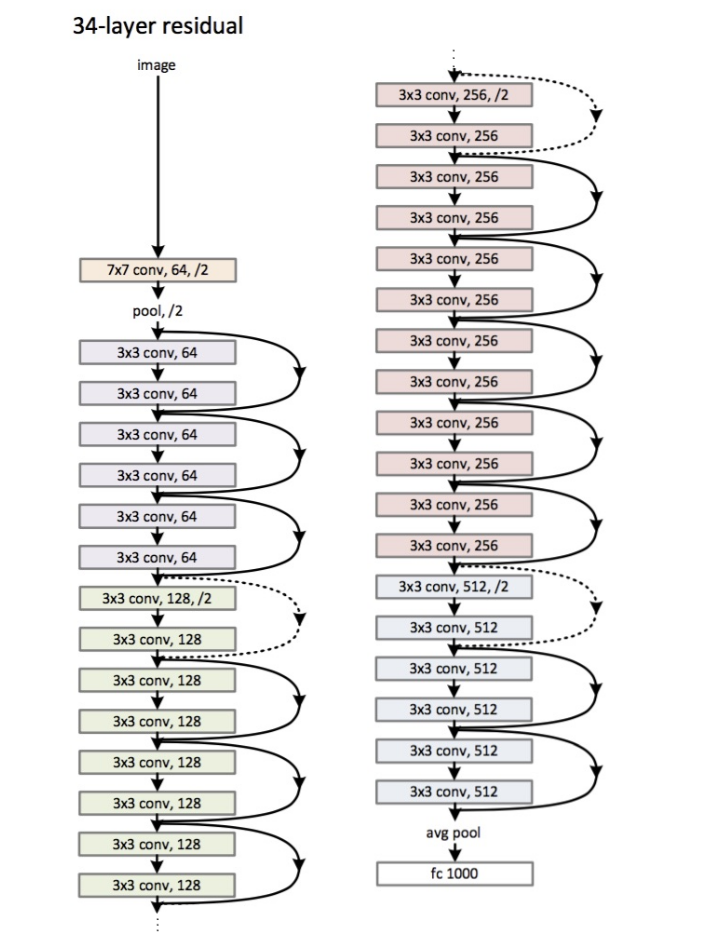}}
  \caption{A 34-layer Residual Convolutional Neural Network described by He et al \cite{he2016deep}. In the transfer learning technique, the last fully connected layer (fc:1000) is removed and a new layer is added with the new number of classes. In the case of network fine-tuning, all the layers are trained with the new dataset while the feature extractor strategy trains only the last classification layer.}
  \label{fig:TL}
\end{figure}

\section{Methodology}
\subsection{Transfer Learning}
Transfer learning takes a different approach to solving
image classification problems in that it utilizes the idea
that many existing architectures and fully trained networks can generalize well to similar problems. This is
possible because images share many underlying features
that are learned in the intermediate layers of the network,
for example edges, corners, and partial objects. In contrast to most neural architecture search implementations,
where once the architecture is discovered the network
weights have to be learned from scratch, transfer learning aims to reuse an existing architecture to either fine-tune the weights on a previously trained network or train
only a few layers, for example, the final classification
layer. This idea has been shown to work when utilizing networks pre-trained on a large image dataset such
as ImageNet \cite{garcia2018out}\cite{garcia2018behavior}, essentially bypassing the most resource-intensive part of training a classifier for a new task
using the full dataset. Utilizing this resource-efficient
technique has also been shown to be successful on
an end-to-end deep learning benchmark competition designed by Stanford University \cite{coleman2017dawnbench}. 

Two main strategies for performing transfer learning
with Convolutional Neural Networks (CNNs) are - 1) network fine-tuning, also referred to as supervised domain adaptation \cite{tajbakhsh2016convolutional},
and 2) convolutional feature extraction \cite{orenstein2017transfer}. To illustrate the strategies, consider the convolutional architecture presented by He et al \cite{he2016deep} shown in Fig. \ref{fig:TL}. They showed that architectures with more layers achieve better accuracy on image classification tasks. Fig. \ref{fig:TL} shows one of their
architectures with 34 layers. Implementations of these
architectures exist for many of the most popular neural
network libraries, such as TensorFlow and PyTorch, in
both randomized weight initializations and fully trained
on datasets such as CIFAR and ImageNet. In both of the aforementioned transfer learning strategies, fully trained
networks are utilized where the last fully connected layer
(fc:1000), as shown in Fig. \ref{fig:TL}, is replaced with a new
fully connected layer corresponding to the number of
classes of the new classifier. The main difference in the
two strategies is the number of layers we finally train in
the new model. In fine-tuning, we use the new dataset
to train the entire network, while in the feature extractor
strategy we freeze most layers, typically all, with the exception of the final fully connected layer (FC), then train
the unfrozen parts. Freezing a layer means forcing it to
not compute the gradients and backpropagating the error
during training, thus not updating the corresponding network
weights. In both strategies, the intuition is that the large
amount of information encoded in the intermediate layers can be efficiently transferred to the new classification
task.

\subsection{Explainability of Deep Learning Models}
In the last few years, there have been significant improvements in the performance of deep learning models. These successful
AI models possess non-linear structure\footnote{The transformation of input to output is complex and cannot be expressed as a linear function.} and behave like black-box systems i.e., the reasons behind their prediction/decisions are unknown to the users. This is a big disadvantage as it leads to a lack of transparency. In order to overcome this issue, research on the explainability 
of deep learning models has drawn attention. The goal of explainability is to measure the importance of the input features and interpret the prediction of an AI model. 

%\vspace{-4mm}
\begin{figure*} [t]
    \centering
    \resizebox{16cm}{!}{%
    \begin{tabular}{c c c}
    \begin{tikzpicture}
\begin{axis}[
    title={Inception v3 (27.1M)},
    xlabel={epochs},
    ylabel={Accuracy},
    ymajorgrids=true,
    grid style=dashed,
    legend style={at={(0.45,-0.25)},
	anchor=north,legend columns=-1},
]
\addplot[color=brown, mark=x] table [x=a, y=b, col sep=comma] {data/INCEPTIONex.csv};
\addplot[color=blue, mark=square] table [x=a, y=b, col sep=comma] {data/INCEPTION.csv};
\addplot[color=red, mark=o] table [x=a, y=b, col sep=comma] {data/INCEPTIONb.csv};
\addplot [mark=*,mark options={scale=0.7,solid}] coordinates {
(0,0.6923) (30,0.6923)
};
\addplot [dashed,mark=*,mark options={scale=0.7,solid}] coordinates {
(0,0.5231) (30,0.5231)
};
\addplot [color=green!60!black, mark=triangle,mark options={scale=1.5,solid}] coordinates {
(0,0.9565) (30,0.9565)
};
\end{axis}
\end{tikzpicture}
         &  
\begin{tikzpicture}
\begin{axis}[
    title={DenseNet161 (28.6M)},
    xlabel={epochs},
    ymajorgrids=true,
    grid style=dashed,
    legend style={at={(0.45,-0.25)},
	anchor=north,legend columns=-1},
]
\addplot[color=brown, mark=x] table [x=a, y=b, col sep=comma] {data/DENSENETex.csv};
\addplot[color=blue, mark=square] table [x=a, y=b, col sep=comma] {data/DENSENET.csv};
%\addlegendentry{cfe}
\addplot[color=red, mark=o] table [x=a, y=b, col sep=comma] {data/DENSENETb.csv};
\addplot [mark=*,mark options={scale=0.7,solid}] coordinates {
(0,0.83) (30,0.83)
};
\addplot [dashed,mark=*,mark options={scale=0.7,solid}] coordinates {
(0,0.58) (30,0.58)
};
\addplot [color=green!60!black, mark=triangle,mark options={scale=1.5,solid}] coordinates {
(0,0.9783) (30,0.9783)
};
\end{axis}
\end{tikzpicture}
    &
\begin{tikzpicture}
\begin{axis}[
    title={ResNet 152 (60.2M)},
    xlabel={epochs},
    ymajorgrids=true,
    grid style=dashed,
    legend style={at={(0.45,-0.25)},
	anchor=north,legend columns=-1},
]
\addplot[color=brown, mark=x] table [x=a, y=b, col sep=comma] {data/RESNET152ex.csv};
\addplot[color=blue, mark=square] table [x=a, y=b, col sep=comma] {data/RESNET.csv};
%\addlegendentry{cfe}

\addplot[color=red, mark=o] table [x=a, y=b, col sep=comma] {data/RESNETb.csv};
%\addlegendentry{rin}
\addplot [mark=*,mark options={scale=0.7,solid}] coordinates {
(0,0.7692) (30,0.7692)
};
\addplot [dashed,mark=*,mark options={scale=0.7,solid}] coordinates {
(0,0.6) (30,0.6)
};
\addplot [color=green!60!black, mark=triangle,mark options={scale=1.5,solid}] coordinates {
(0,0.9565) (30,0.9565)
};
\end{axis}
\end{tikzpicture}

\\ % SECOND ROW

 \begin{tikzpicture}
\begin{axis}[
    title={ResNet 18 (11.7M)},
    xlabel={epochs},
    ylabel={Accuracy},
    ymajorgrids=true,
    grid style=dashed,
]
\addplot[color=brown, mark=x] table [x=a, y=b, col sep=comma] {data/RESNET18ex.csv};
\addplot[color=blue, mark=square] table [x=a, y=b, col sep=comma] {data/RESNET18.csv};

\addplot[color=red, mark=o] table [x=a, y=b, col sep=comma] {data/RESNET18b.csv};

\addplot [mark=*,mark options={scale=0.7,solid}] coordinates {
(0,0.6923) (30,0.6923)
};
\addplot [dashed,mark=*,mark options={scale=0.7,solid}] coordinates {
(0,0.6) (30,0.6)
};
\addplot [color=green!60!black, mark=triangle,mark options={scale=1.5,solid}] coordinates {
(0,0.8913) (30,0.8913)
};

\end{axis}
\end{tikzpicture}
         &  
\begin{tikzpicture}
\begin{axis}[
    title={VGG11 BN (132.8M)},
    xlabel={epochs},
    ymajorgrids=true,
    grid style=dashed,
    legend style={at={(0.45,-0.25)},
	anchor=north,legend columns=-1},
]
\addplot[color=brown, mark=x] table [x=a, y=b, col sep=comma] {data/VGGex.csv};
\addplot[color=blue, mark=square] table [x=a, y=b, col sep=comma] {data/VGG.csv};
\addplot[color=red, mark=o] table [x=a, y=b, col sep=comma] {data/VGGb.csv};
\addplot [mark=*,mark options={scale=0.7,solid}] coordinates {
(0,0.8308) (30,0.8308)
};
\addplot [dashed,mark=*,mark options={scale=0.7,solid}] coordinates {
(0,0.6615) (30,0.6615)
};
\addplot [color=green!60!black, mark=triangle,mark options={scale=1.5,solid}] coordinates {
(0,0.978) (30,0.978)
};

\end{axis}
\end{tikzpicture}
    &
\begin{tikzpicture}
\begin{axis}[
    title={SqueezeNet v1 (1.2M)},
    xlabel={epochs},
    ymajorgrids=true,
    grid style=dashed,
    legend style={at={(0.45,-0.25)},
	anchor=north,legend columns=-1},
]
\addplot[color=brown, mark=x] table [x=a, y=b, col sep=comma] {data/SQNETex.csv};
\addplot[color=blue, mark=square] table [x=a, y=b, col sep=comma] {data/SQNET.csv};
\addplot[color=red, mark=o] table [x=a, y=b, col sep=comma] {data/SQNETb.csv};
\addplot [mark=*,mark options={scale=0.7,solid}] coordinates {
(0,0.8) (30,0.8)
};
\addplot [dashed,mark=*,mark options={scale=0.7,solid}] coordinates {
(0,0.4769) (30,0.4769)
};
\addplot [color=green!60!black, mark=triangle,mark options={scale=1.5,solid}] coordinates {
(0,0.8261) (30,0.8261)
};
\end{axis}
\end{tikzpicture}

\end{tabular}
}
\caption{Models trained on the operational TEL dataset. \textbf{Square} markers show validation accuracy of models trained with transfer learning via full model fine-tuning (all model parameters). \textbf{Circle} markers represent networks trained on the same dataset without transfer from random initialization. \textbf{X} markers show validation accuracy of models trained with transfer learning by retraining only the last classification layer of the model to examine raw feature reuse. The approximate number of trainable parameters is specified next to architecture names in millions. \textbf{Horizontal line with triangle} markers shows test set accuracy of the best model selected over 30 epochs. \textbf{Horizontal solid line with circle} markers show the accuracy of the best transfer model on the OOD set, and the \textbf{dashed line} shows the accuracy of the best model without transfer on the OOD set.
Implemented architectures were ~\cite{RESNET:7780459,inception:7780677,DenseNet:8099726,DBLP:journals/corr/SimonyanZ14a,squeeze:DBLP:journals/corr/IandolaMAHDK16}.
All training runs reported were performed with $\alpha$ of 0.001, stochastic gradient descent with momentum of 0.9 as the optimizer, step size of 7, and a $\gamma$ value of 0.1. Input dimensions were 224x224 and 299x299 for Inception. 
}       
    \label{tab:transfer_learning}
\vspace{-4mm}
\end{figure*}

There are two main kinds of explainability methods - gradient-based and perturbation-based. In the case of gradient-based algorithms, gradients are used to compute the importance between input features and the output \cite{selvaraju2017grad}. In the case of perturbation-based methods, values very close to the input are taken into consideration and the difference in the output is calculated in order to determine the importance of input features \cite{ribeiro2016should}. 

According to \cite{gilpin2018explaining}, an explanation can be based on its interpretability or its completeness. In the case of interpretability, the target is to make the internals of the system understandable to the general users. In the case of a complete explanation, the focus is mainly on the accuracy. For example, presenting all the mathematical operations of a deep learning model is a complete explanation. There are different techniques to generate explanations of deep learning models. Firstly, the focus is on the data processing by the model, i.e. why an output is generated from a set of inputs. Both gradient-based and perturbation-based methods are implemented using this technique. Secondly, the focus is on the information of the model. Also, explanation-generating systems can be created in order to generate explanations understandable to general users \cite{gilpin2018explaining}.

Our focus is mainly on gradient-based explainability methods. Class Activation Mapping \cite{zhou2016learning}, Gradient Weighted Class Activation Mapping \cite{selvaraju2017grad} etc. are robust methods for generating visual explanations. In \cite{zhou2016learning}, Class Activation Maps were used for visually explaining CNNs. In CNN, global average pooling is used to generate class activation maps (CAMs). For a specific class, a class activation map highlights the discriminative regions in the image that the CNN uses to predict that class. In order to generate CAMs, the predicted class score is mapped back to the previous convolutional layer. It was observed that for a given image, the class-specific discriminative regions were different for different classes \cite{zhou2016learning}. In \cite{selvaraju2017grad}, the Gradient-Weighted Class Activation Mapping (Grad-CAM) technique was proposed to generate visual explanations of the decisions of CNN models. In this technique, the gradients of the target class are used, flowing into the last convolutional layer to generate a class-discriminative localization map for identifying the target class. No architectural changes or re-training were needed for producing visual explanations of CNN models using the Grad-CAM technique. 

\begin{table*} [t]
\vspace{-4mm}
      \centering
      \caption{HEATMAPS OF TEST IMAGES FOR VGG11 ARCHITECTURE. The irrelevant areas in the heatmaps convert from green, yellow, and red (in column two) to blue and purple (in column three). The opposite case occurs for relevant TEL areas. This visually explains the better performance of the CNN architectures when transfer learning is used.} 
      \begin{tabular}{|m{1cm}|m{4cm}|m{4cm}|m{4cm}|}
        \hline       
        {\textbf{No.}} & {\textbf{Test Image}} & {\textbf{Pre-trained VGG11 on ImageNet}} & {\textbf{Trained VGG11 on TEL Dataset with Transfer Learning}} \\
        \hline       
        {1} & {\includegraphics[width=4cm, height=3cm]{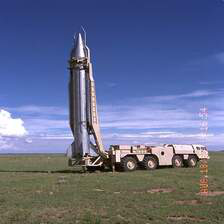}} & {\includegraphics[width=4cm, height=3cm]{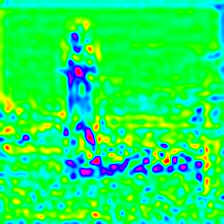}} & {\includegraphics[width=4cm, height=3cm]{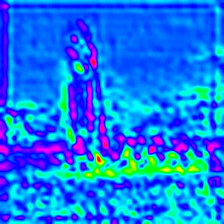}} \\
        \hline
        {2} & {\includegraphics[width=4cm, height=3cm]{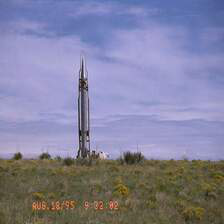}} & {\includegraphics[width=4cm, height=3cm]{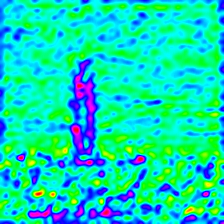}} & {\includegraphics[width=4cm, height=3cm]{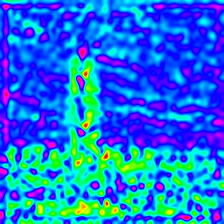}} \\
        \hline
         {3} & {\includegraphics[width=4cm, height=3cm]{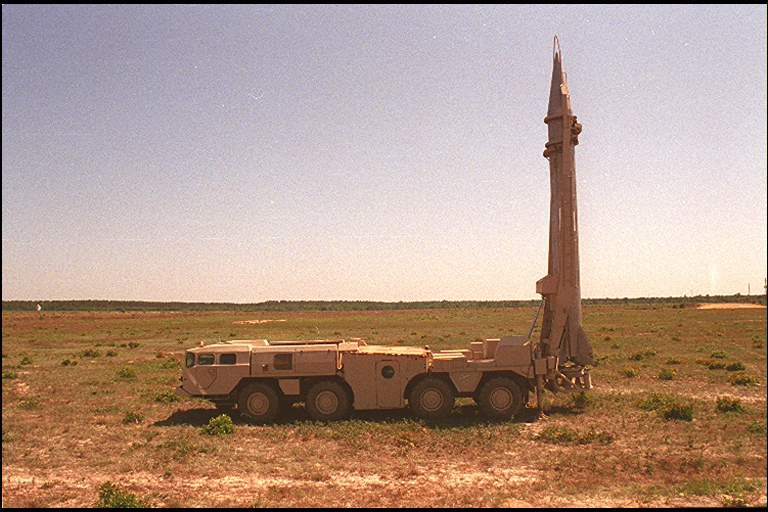}} & {\includegraphics[width=4cm, height=3cm]{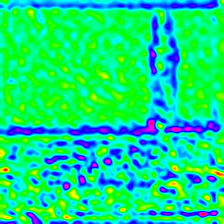}} & {\includegraphics[width=4cm, height=3cm]{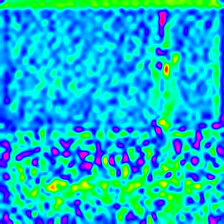}} \\
        \hline 
        {4} & {\includegraphics[width=4cm, height=3cm]{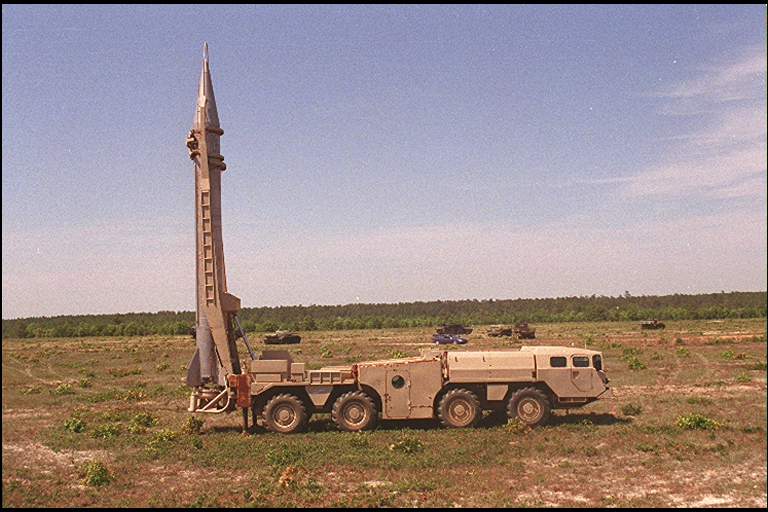}} & {\includegraphics[width=4cm, height=3cm]{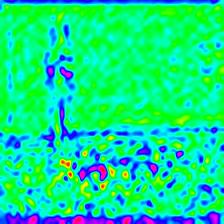}} & {\includegraphics[width=4cm, height=3cm]{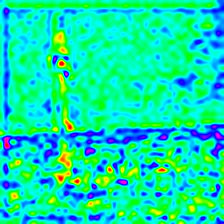}} \\
        \hline            
     \end{tabular}      
     \label{tab:heatmap}
\vspace{-4mm}
\end{table*}

%title: Data and Models Regime
\section{Application Domain and its Data Regime}\label{sec:experiment}  
Our application domain pertains to the classification of images that may contain specific types of military hardware.
For this analysis, we focused on a specialized hardware called transporter erector launcher (TEL), a type of military hardware used to transport and launch missiles, as a target class;
our goal was to train classifiers to detect the presence of TELs in images. We are restricted to a small number of labeled samples due to the novelty of the class.
Our labeled data set contained 100 images of TELs and 130 images where there was no TEL, which we split into \emph{train/val/test} sets using a 60/20/20 ratio. We also collected a set of open-source images as an OOD set used to evaluate the effectiveness of TL under possible covariate shift\footnote{Covariate shift means the distribution of input data shifts between the training environment and live environment.}. The OOD set contained 31 images of TELs and 34 images where there was no TEL.
%Images were resized to 224x224 pixels (299x229 for Inception) and normalized using ImageNet's  $\mu$ and $\sigma$.

\section{Experimental Setup}
Six CNN models implemented in PyTorch and trained on ImageNet were used in this experiment (see Fig.~\ref{tab:transfer_learning}). The CNN models were - Inception v3, Densenet161, ResNet 152, ResNet 18, VGG11 and SqueezeNet v1. First, the data directory was specified. Then the architecture was selected and the final classification layer was set for two classes.
It was selected whether whole model fine-tuning or feature extraction would be done by setting the Freeze flag.
A model name was specified to save the model.
Then the code was run for retraining/fine-tuning a new model based on our data. For 30 epochs, the model training and selection process iterated. The best performing model was saved based on accuracy. The models trained on the original TEL dataset were tested on the test images of the TEL dataset and OOD dataset\footnote{Out-of-distribution (OOD) data differs from training data, thus can be used to test the generalization performance of the model.}. Three cases were considered during the experiment:
1) models were trained with transfer learning via full model fine-tuning (all model parameters), 2) models were trained without transfer learning, and 3) models were trained with transfer learning by retraining only the last classification layer. We note that the reported results had no hyper-parameter optimization or K-fold validation, as our goal was to investigate the relationship of feature reuse and overparameterization in data-starved regimes.

\begin{table*} [t]
\vspace{-4mm}
      \caption{VALIDATION (ON TEL DATASET) AND TEST (ON OOD DATASET) ACCURACY OF CNN MODELS WITH AND WITHOUT TRANSFER LEARNING}
      \label{tab:accuracy}
      \centering
      \renewcommand{\arraystretch}{1.2}
      \begin{tabular}{|m{2cm}|m{2.5cm}|m{2.2cm}|m{1.6cm}|m{1.8cm}|m{1.8cm}|}
        \hline       
        {\textbf{Model}} & {\textbf{Validation Accuracy with Transfer (Full Model Fine-Tuning)}} & {\textbf{Validation Accuracy with Transfer(without Fine-Tuning)}} & {\textbf{Validation Accuracy without Transfer Learning}} & {\textbf{Test Accuracy on OOD set with Transfer}} & {\textbf{Test Accuracy on OOD set without Transfer}}\\
        \hline 
        {Inception v3} & {98-100\%} & {87-89\%} & {70-75\%} & {70\%} & {52\%}\\
        \hline
        {DenseNet161} & {100\%} & {94-96\%} & {85-89\%} & {83\%} & {58\%}\\
        \hline
        {ResNet 152} & {100\%} & {88-92\%} & {65-72\%} & {78\%} & {60\%}\\
        \hline
        {ResNet 18} & {98\%} & {87-89\%} & {80-87\%} & {79\%} & {60\%}\\
        \hline
        {VGG11 BN} & {96-98\%} & {87-89\%} & {70-80\%} & {83\%} & {66\%}\\
        \hline
        {SqueezeNet v1} & {90\%} & {88-90\%} & {44\%} & {80\%} & {48\%}\\
        \hline         
     \end{tabular}
\vspace{-4mm}
\end{table*}  

We used heatmaps to visualize features captured by the transfer-learned CNN architecture. We adopted the Gradient-weighted Class Activation Mapping technique to generate heatmaps \cite{selvaraju2017grad}. Specifically, we applied this technique to four randomly selected test images from our TEL dataset, where each image contains at least one TEL. The heatmaps were generated using the feature information from the 11th layer of the VGG11 architecture. Table \ref{tab:heatmap} shows the generated heatmaps of four random test images which highlight the regions responsible for the detection of TELs in the test images. We chose VGG11 as the test model because it exhibited stable performance and achieved the highest accuracy on the test set.

\section{Result \& Discussion}
As shown in Figure~\ref{tab:transfer_learning}, all tested architectures enhanced their accuracy performance by utilizing transfer learning on both the operational TEL dataset and the OOD dataset. All the validation and test accuracy are shown in Table \ref{tab:accuracy}. In the case of Inception v3 and Resnet 152 models, the validation accuracy without transfer was below 80\% whereas the validation accuracy with transfer was above 80\%, 100\% validation accuracy for Resnet 152 model trained with transfer learning via full model fine-tuning. Also, the DenseNet161 model achieved 100\% validation accuracy with transfer learning, without transfer the validation accuracy was below 90\%. In the case of SqueezeNet v1, the validation accuracy was around 90\% with transfer and without transfer, the validation accuracy was around 44\% (almost half). In the case of the DenseNet161 model, there was a significant difference in the accuracy on the OOD set with transfer and without transfer - with transfer, the accuracy was above 80\% whereas without transfer the accuracy was below 60\%. The difference was even more for SqueezeNet v1 architecture - 80\% accuracy on the OOD set with transfer, the accuracy was below 50\% without transfer. These numbers prove the effectiveness of transfer learning in data-starved environments.

The training pipeline was executed on a computer equipped with an NVIDIA RTX 2080 GPU, PyTorch as the ML toolkit, and Ubuntu 18.04.3. 
Regarding the effect of overparametrization, we did not observe a significant improvement in OOD test set accuracy
between two extremes in terms of the number of trainable parameters,
specifically, SqueezeNet and VGG11, where the latter contains 100x more trainable parameters.

We posit that the significant benefit of transfer in our operational data is associated with the fact that both source and target domains include natural images, hence feature reuse happens at more layers compared to transfer observed in the medical imaging study~\cite{raghu2019transfusion}.
This hypothesis is also supported by limiting retraining to only the last classification layer of the network,
yielding better performance than training without transfer and similar performance as fine-tuning the entire network on the SqueezeNet architecture.

In Table \ref{tab:heatmap}, the first column shows the test images, and the second and third columns show the heatmaps generated using the pre-trained VGG11 model (on ImageNet) and trained VGG11 model (on TEL dataset with transfer learning) respectively. The red, yellow, and green regions indicate the most relevant areas in the heatmaps while the darker regions (blue and purple) indicate the least relevant areas \cite{chaudhuri2021attention}. The heatmaps of the second column show that most of the relevant TEL areas are blue and purple while the irrelevant areas are green, yellow, and red. The heatmaps of the third column show that most of the irrelevant areas are blue and purple while the relevant TEL areas are mostly red, yellow, and green. This indicates a clear improvement in the recognition performance of the CNN architectures when the transfer learning technique is implemented. 

\section{Conclusion}
We conclude that transfer learning with overparameterized convolutional neural networks in data-starved regimes is beneficial and practical, provided that the source and target domains are similar as is the case of natural images represented in ImageNet and our operational data. In our experiments, the accuracy of all CNN architectures improved by using transfer learning. The improved performance of the CNN architectures while utilizing transfer learning is backed up by our CAM-based visual explanation. Our experimental results prove the effectiveness of the transfer learning technique in data-starved environments.

{\footnotesize \bibliographystyle{unsrt}
 \bibliography{refs}}

%% \bibliographystyle{plain}
%% \bibliography{capi}

%\clearpage

\end{document}